\documentclass[sigconf]{acmart}

\renewcommand\footnotetextcopyrightpermission[1]{} 
\usepackage{hyperref}
\usepackage{url}
\usepackage{enumitem}
\usepackage[normalem]{ulem}
\usepackage{algorithm}
\usepackage{algorithmic}
\usepackage{amsmath}
\usepackage{booktabs}

\usepackage{colortbl}
\usepackage{multirow}
\usepackage{makecell}
\usepackage{xcolor}
\usepackage{rotating} 
\usepackage{graphicx}
\definecolor{bestcolor}{RGB}{240, 248, 255}  
\definecolor{improvecolor}{RGB}{255, 250, 240} 
\definecolor{categorycolor}{RGB}{245, 245, 245} 

\usepackage{wrapfig}
\usepackage{subfig}
\usepackage{float}
\usepackage{array}
\usepackage{caption}
\usepackage{tcolorbox}
\usepackage[mathscr]{euscript}
\theoremstyle{definition}
\newtheorem{definition}{Definition}[section]

\newtheorem{proposition}{Proposition}[section]

\AtBeginDocument{%
  }

\setcopyright{acmlicensed}
\copyrightyear{2018}
\acmYear{2018}
\acmDOI{XXXXXXX.XXXXXXX}
\acmConference[Conference acronym 'XX]{Make sure to enter the correct
  conference title from your rights confirmation email}{June 03--05,
  2018}{Woodstock, NY}
\acmISBN{978-1-4503-XXXX-X/2018/06}

\begin{document}

\title{RxnNano:Training Compact LLMs for Chemical Reaction and Retrosynthesis Prediction via Hierarchical Curriculum Learning}

\author{
	Ran Li$^{1}$, \ Shimin Di$^{3}$, \ Haowei LI$^{1}$, \ Luanshi Bu$^{3}$, \ Jiachuan Wang$^{1}$, Wangze Ni$^{4}$, \ Lei Chen$^{1,2}$}
\affiliation{%
	\institution{$^1$The Hong Kong University of Science and Technology, Hong Kong SAR, China}
	\institution{$^2$The Hong Kong University of Science and Technology (Guangzhou), Guangzhou, Guangdong, China}
	\institution{$^3$Southeast University, Nanjing,\ Jiangsu, China}
	\institution{$^4$Zhejiang University, Hangzhou, Zhejiang, China}
	\country{}
}
\email{rlibb@connect.ust.hk, shimin.di@seu.edu.cn,  hliei@connect.ust.hk, luanshibu@seu.edu.cn}
\email{jwangey@connect.ust.hk, niwangze@zju.edu.cn,  leichen@hkust-gz.edu.cn}

\renewcommand{\shortauthors}{Ran et al.}

\begin{abstract}
Chemical reaction prediction is pivotal for accelerating drug discovery and synthesis planning. Despite advances in data-driven models, current approaches are hindered by an overemphasis on parameter and dataset scaling. Some methods coupled with evaluation techniques that bypass fundamental challenges in reaction representation and fail to capture deep chemical intuition like reaction common sense and {topological atom mapping logic}. We argue that the core challenge lies in instilling these knowledge into the models. To this end, we propose a unified framework that prioritizes chemical understanding over scale through three key innovations: (1) a {Latent Chemical Consistency} objective that models reactions as movements on a continuous chemical manifold, ensuring reversible and physically plausible transformations; (2) a {Hierarchical Cognitive Curriculum} that trains the model through progressive stages, from syntax mastery to semantic reasoning, building robust chemical intuition; (3) {Atom-Map Permutation Invariance (AMPI)}, which force the model to learn invariant relational topology and balance multi-task learning. (4)and structured plan-based reasoning to improve the performance of the LLMs. Our compact {0.5B-parameter model}, \textbf{RxnNano} significantly outperforms fine-tuned LLMs ten times larger (>7B) and all the domain baselines, achieving a 23.5\% Top-1 accuracy improvement on rigorous benchmarks without test-time augmentation. \url{https://github.com/rlisml/RxnNano}.
\end{abstract}

%

\keywords{Retrosynthesis, AI4Science, Large Language Model}


\maketitle

\section{Introduction}

Chemical reaction prediction stands as a cornerstone challenge in computational chemistry, serving as the engine for accelerated drug discovery and automated synthesis planning~\citep{zhao2025comprehensive,long2025artificial,zhu2023dual,xie2022retrograph}. This domain encompasses two complementary tasks: \textit{forward reaction prediction}~\citep{schwaller2019molecular}, which infers products from reactants, and \textit{retrosynthetic analysis}~\citep{ zhao2025comprehensive}, which reasons backward from a target molecule to identify plausible reactants. While traditional methods relied on expert-curated rules~\citep{szymkuc2016computer,bogevig2015route}, the field has shifted toward data-driven paradigms~\citep{wang2021retroprime,coley2017prediction,segler2017neural} that learn from large-scale reaction databases~\citep{jacob2018statistics,kim2025pubchem}.

Chemical data is fundamentally represented either as molecular graphs (nodes and edges) or linearized strings, most notably simplified molecular input line entry system (SMILES). Consequently, modeling approaches generally align with these modalities. \textit{Graph-based models}~\citep{chen2019graph,dai2019retrosynthesis,chen2023g} utilize Graph Neural Networks (GNNs) to capture topological structures, while \textit{Sequence-based models}~\citep{liu2017retrosynthetic,karpov2019transformer,zheng2019predicting}  treat reaction prediction as a translation task using Transformers. Recently, \textit{Large Language Models (LLMs)}~\citep{zhao2025developing,zhao2024chemdfm,yu2024llasmol,ramos2025review} have emerged as a potential powerful solutions to predict reactions. However, without continue training, even models like O1~\citep{jaech2024openai} mini and DeepSeek-R1~\citep{guo2025deepseek} can underperform the domain models.
Functionally, these methods fall into three categories: (1) \textit{Template-based methods}~\citep{dai2019retrosynthesis,chen2021deep,xie2023retrosynthesis} , which apply rigid, predefined reaction rules but fail to generalize to novel chemistry; (2) \textit{Semi-template methods}~\citep{somnath2021learning,zhong2023retrosynthesis}, which offer more flexibility but struggle with complex multi-step transformations; and (3) \textit{Template-free methods}~\citep{wan2022retroformer,han2024retrosynthesis}, which directly generate product structures. While template-free approaches offer the highest generality, they are prone to generating chemically invalid molecules~\citep{schwaller2018found}. It would be ideal to combine the benefit of both template-based and template-free methods.

\begin{figure}[!t]
	\centering
	\includegraphics[width=0.49\textwidth]{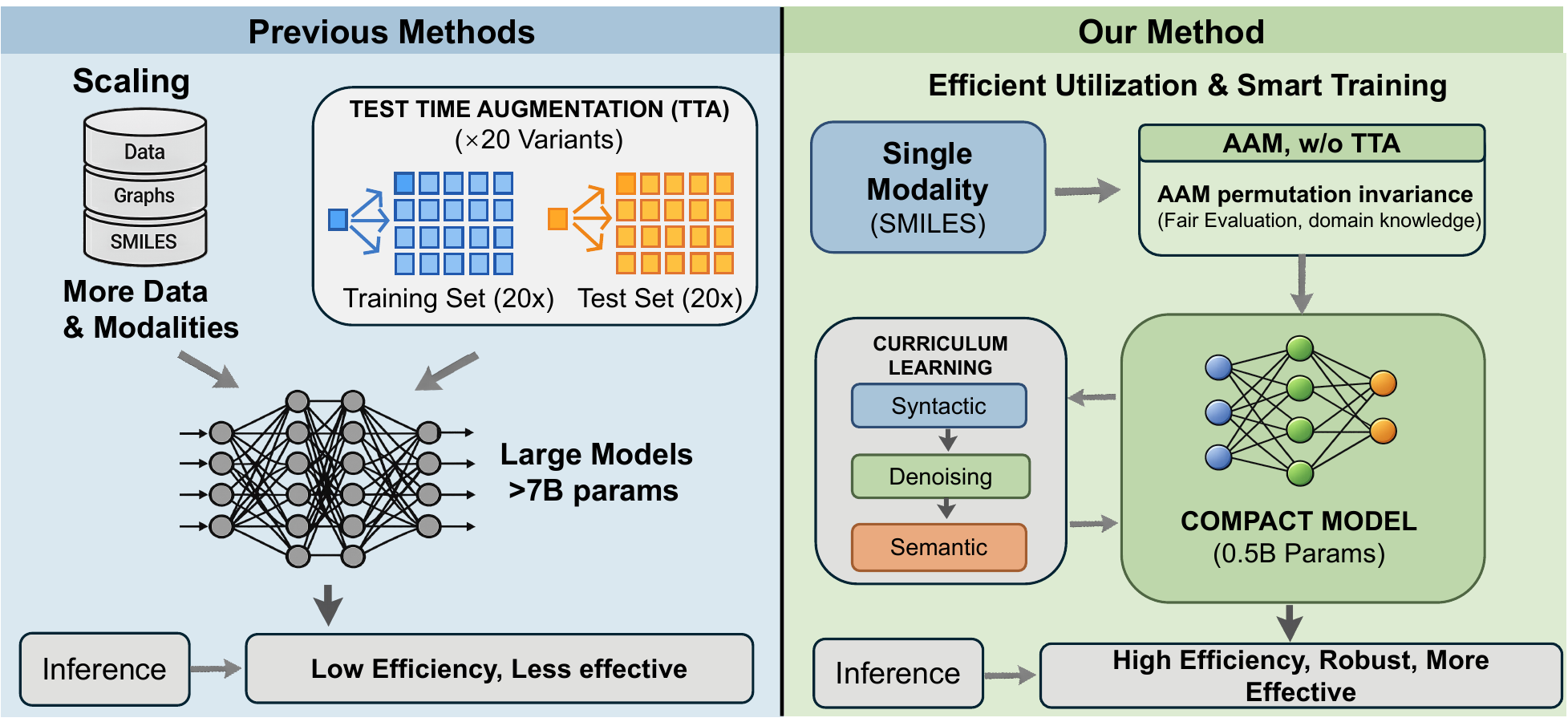}
	\caption{Comparison of ours and baselines}
	\label{fig:motivation}
\end{figure}

Despite the proliferation of models, progress is currently impeded by a focus on parameter scaling and evaluation tricks rather than deep chemical understanding as shown in Figure~\ref{fig:motivation}. We identify three critical issues in the current literature:
\textit{1.Inefficient Scaling and Modality Noise:} The prevailing trend favors increasing model size and merging diverse data modalities under the assumption of scaling laws. However, without refined data processing, integrating multimodal data into larger models often introduces modality noise rather than meaningful signal. Large-scale models~\citep{taylor2022galactica,zhao2025developing,zhu2023dual} can achieve suboptimal accuracy (e.g., 17.9\% top-1 on USPTO-50k) despite massive parameter counts, suggesting that scale alone cannot compensate for a lack of domain-specific inductive bias.
\textit{2.Evaluation on augmented data:} 
Current literature exhibits critical inconsistencies in evaluation methodology. Many models~\citep{han2024retrosynthesis,zhang2025reasoning,yan2020retroxpert,wang2021retroprime,seo2021gta,tetko2020state,irwin2022chemformer,zeng2024ualign,wang2023retrosynthesis} employ extensive test-time augmentation (e.g., 20× or more) on both training and test sets, creating synthetic evaluation scenarios where models are tested on heavily augmented data rather than genuine chemical examples.     
\textit{3.Misuse of Atom-Atom Mapping (AAM):} AAM provides high-quality signals regarding atomic correspondence during reactions~\citep{maziarz2025re}. It assigns a unique number to each atom across the reaction. 	
However, the current literature lacks a unified standard for its usage. Some models leverage AAM to guide training but are compared against baselines that do not use it, creating an unfair comparison. Moreover, its direct use might result in over-reliance on atom mapping numbers for prediction, hindering generalization to scenarios without AAM information.  Others discard this rich information entirely, resulting in ineffective learning. AAM can provide high quality training signal and a new training strategy should be adopted to learn from this process while preventing the models from relying on the mapping number only for prediction~\citep{xu2025unified}.

We argue that the core challenge of chemical AI is not merely scaling, 
but the incorporation of \textit{reaction common sense} and \textit{topological atom–to–atom mapping logic} into the model's architectural and learning process. Rather than treating SMILES strings as arbitrary sequences, a model must understand the underlying manifold of chemical transformations. 
We argue that by focusing on data granularity and training depth, it is possible to construct a compact model that outperforms larger models without TTA.
This indicates that the path to better reaction prediction lies in first advancing our understanding of the data itself and designing a more principled, fine-grained training paradigm, prior to further scaling. Moreover, 
it is also important to ground the LLM's output in this areas. Previous methods tends to distill knowledge from a larger LLMs~\citep{zhang2025reasoning}. However, since the performance of LLMs without further training on this tasks is bad, distillation from such a model can result in noise rather than consistent improvement.

\begin{figure*}[!t]
	\centering
	\includegraphics[width=0.95\textwidth]{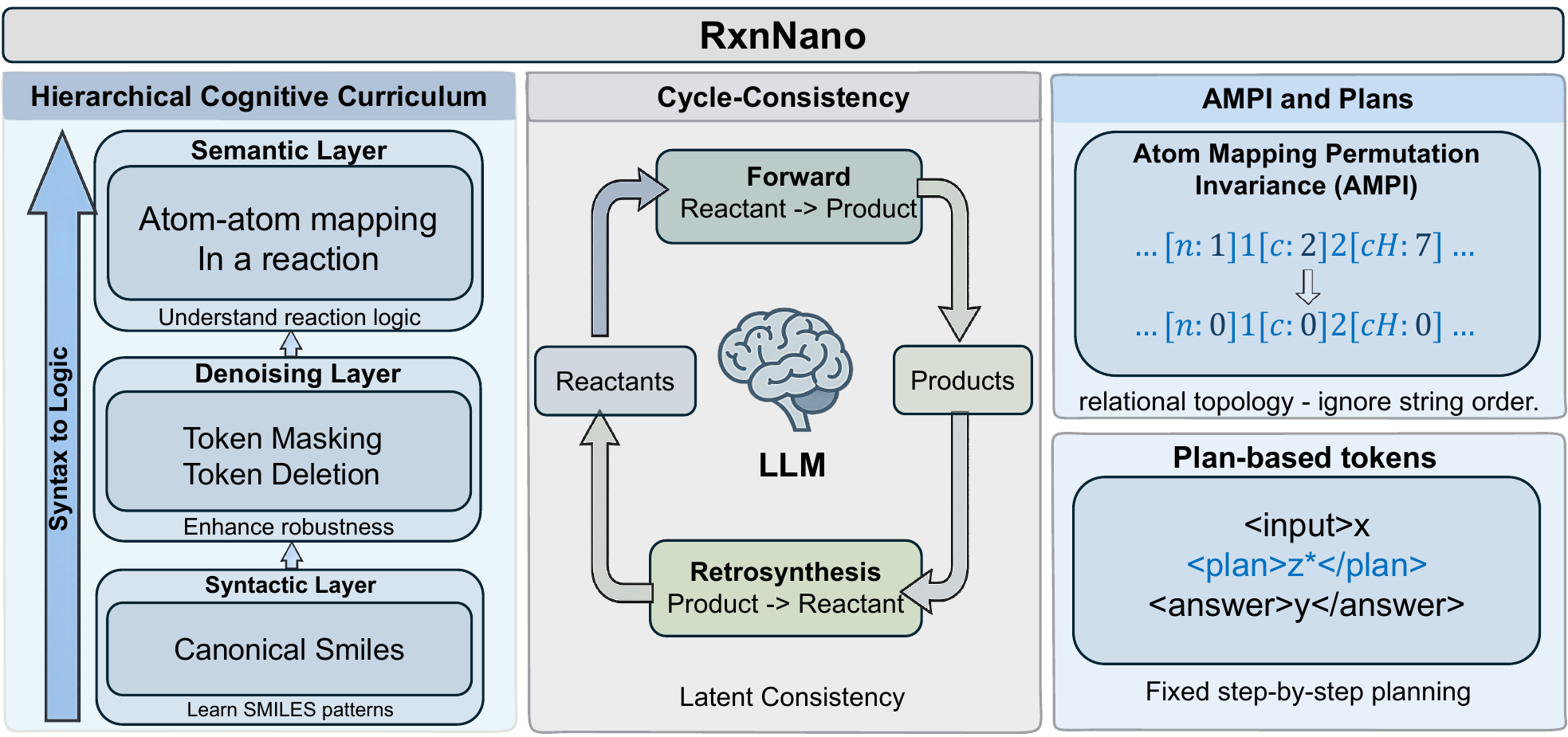}
	\caption{Overview of our framework: The Hierarchical Curriculum from syntax to semantics. The Cycle-Consistency ensuring topological robustness. The Relational Invariance mechanism for unbiased chemical reasoning and Plan-based token to enhance the inference.}
	\label{fig:overview}
\end{figure*}

To achieve  such a paradigm, we propose a unified framework as shown in Figure~\ref{fig:overview} that prioritizes deep chemical understanding over
 scaling.
Our approach is built upon three fundamental pillars.
We move beyond 
just sequence mapping
by treating reactions as trajectories within a continuous chemical manifold. 
We introduce a \textbf{Latent Cycle Consistency} objective, enforcing that the latent representation of a molecule remains invariant when cycled through forward and retrosynthetic logic. 
This ensures the model captures the underlying physics of transformation rather than just memorizing token transitions.
structured "Grammar to Logic" progression.
We posit that robust reaction knowledge should be acquired through a structured "From Grammar to Logic" progression.
Our model follows a \textbf{Hierarchical Curriculum Learning} progression: starting with a \textit{Syntactic phase} to master SMILES syntax and distributions, moving to a \textit{Denoising phase} to repair structural errors, and culminating in a \textit{Semantic phase} that learns the underlying atom-to-atom correspondence in a reaction. This curriculum builds a foundation of syntactic robustness before tackling complex reaction mechanisms.
In this way, the model develops reaction common sense by deep mining one modality well and reducing the need for brute-force parameter scaling in the current tasks.
To prevent the model from exploiting dataset biases (e.g., specific ordering of atoms), we implement \textbf{Atom-Map Permutation Invariance (AMPI)}. This encourages the model to learn the Relational Topology, the atom-to-atom mapping across the reaction, rather than superficial string patterns or indices. This ensures fairness in comparison (models don't get an unfair advantage from index hints) and guarantees generalization to real-world scenarios without perfect AAM. Based on this, we trained two versions of models one with full data information including AAM. Another is trained and evaluated without the AAM numbers.
To improve the performance of LLMs without relying on distillation from unreliable large LLMs, we propose
\textbf{Structured plan-based reasoning} that uses fixed explicit step-by-step rationalizations, enhance the training without requiring annotated chain-of-thought data. 

Our framework demonstrates that strategic architectural design can supersede scaling. We show that our compact {0.5B parameter model}, \textbf{RxnNano}, when equipped with these strategies, 
significantly outperforms 7B  parameter LLMs and other domain models. Specifically, our model achieves a +23.5\% accuracy improvement over strong baselines while maintaining high computational efficiency for downstream planning. 
It is noteworthy that our model, evaluated without an AAM, still delivers consistent performance gains over the baselines which utilize AAM.
By resolving the evaluation discrepancy between mapped and unmapped data, we provide a more rigorous and realistic benchmark for the field.

\section{Related Work}

Chemical reaction prediction has transitioned from expert-defined systems to data-driven deep learning architectures. This section reviews the landscape of AI-driven chemistry through three lenses: molecular representation, methodological paradigms, and the technical implementation strategies that enhance model reasoning.

\subsection{Molecular Representations and Chemical Encoding}
The effectiveness of a model is inherently tied to how it "perceives" a molecule~\citep{wigh2022review}. While modern methods utilize diverse modalities and larger datasets, they often overlook the "deep understanding" of the representation itself. Current research primarily utilizes three encoding strategies:

{SMILES Strings (Sequence-based)}: The Simplified Molecular Input Line Entry System (SMILES) represents molecules as linear text strings~\citep{weininger1988smiles}. For example, Ethanol is written as \texttt{CCO}. Because Large Language Models (LLMs) favor sequential data, SMILES have become the predominant choice for generative chemistry. However, they suffer from "fragility"—a single character change can describe a completely different or invalid molecule—requiring models to learn complex syntax alongside chemical rules.
{Molecular Graphs}: These treat atoms as nodes and bonds as edges. While Graph Neural Networks (GNNs) excel at capturing local structural information, they often struggle with long-range dependencies across complex molecular frameworks~\citep{chen2023g,chen2019graph,dai2019retrosynthesis,shi2020graph}.
{Atom-Atom Mapping (AAM)}: AAM acts as a "traceability" layer, indexing each reactant atom to its corresponding position in the product. While AAM provides high-quality supervision, a significant challenge remains: making models learn the underlying \textit{mapping logic} rather than simply memorizing numerical atom indices, which do not generalize to unmapped real-world data.

\subsection{Methodological Paradigms}
Current approaches vary based on the trade-off between explicit chemical "priors" and learned patterns.

\textbf{Template-based Methods}
These rely on a library of predefined chemical transformations. For example, a template might define the specific subgraph change of an esterification reaction. While highly interpretable and chemically accurate~\citep{chen2021deep, xie2023retrosynthesis}, they are limited by the "out-of-vocabulary" problem, that is they cannot predict transformations absent from their training library~\citep{coley2017computer,shee2024site}.

\textbf{Semi-template Methods}
These bridge the gap by first identifying a "reaction center" (the specific atoms undergoing change) and then applying local modifications~\citep{shi2020graph, zhong2023retrosynthesis}. This narrows the search space while maintaining more flexibility than fixed templates, though they remain sensitive to the accuracy of the initial center prediction.
\textit{Template-free methods}~\citep{sacha2021molecule,wan2022retroformer,seo2021gta,han2024retrosynthesis} represent the most flexible approach, treating reaction prediction as a sequence-to-sequence or graph-to-graph translation problem without explicit reaction rules. While offering greater generality and can potentially discover novel reactions, they face significant challenges: tendency to generate chemically implausible products that violate fundamental chemical constraints~\citep{schwaller2018found,han2024retrosynthesis,zeng2024ualign}.
\textbf{LLMs in Chemistry}
The massive success of LLMs in structured domains like mathematics and computer programming has motivated their application to chemistry~\citep{zhao2024chemdfm}. While early general-purpose models like Galactica~\citep{taylor2022galactica} showed limited performance on specialized tasks like retrosynthesis~\citep{zhao2025developing}, they possess a unique potential. Unlike previous paradigms, a properly trained LLM have the potential to integrate diverse information types, offering the flexibility of template-free models while internalizing the rigid logic of template-based systems. Our work builds on this potential, exploring how LLMs can leverage sequential reasoning to master reaction prediction.

\subsection{Further Techniques for Performance Enhancement}
Beyond core architectures, several technical strategies are employed to improve model performance, though their impact on true generalization remains a subject of debate.

\textbf{Test-Time Augmentation (TTA):}
Since a single molecule can be represented by multiple valid SMILES strings (e.g., \texttt{CN} and \texttt{NC} are both Methylamine), TTA involves feeding multiple permutations of the same input into the model~\citep{han2024retrosynthesis,zhang2025reasoning,yan2020retroxpert,wang2021retroprime,seo2021gta,tetko2020state,irwin2022chemformer,zeng2024ualign,wang2023retrosynthesis}. While data augmentation during training generally improves robustness, a concerning trend in current literature is the heavy use of TTA during evaluation (e.g., 20x augmentation of the test set). It has been observed that while metrics may increase on these augmented test sets, performance often drops when the same model is evaluated on the original, non-augmented evaluation set. This suggests that excessive TTA may artificially inflate scores rather than improving the model's fundamental chemical reasoning.

\textbf{Fair Evaluation Protocols:}
A critical issue in the field is the inconsistent use of Atom-Atom Mapping (AAM) and TTA in benchmarking~\citep{maziarz2025re}, which obscures the true capabilities of different models~\citep{han2024retrosynthesis,zhang2025reasoning,yan2020retroxpert,wang2021retroprime,seo2021gta,tetko2020state,irwin2022chemformer,zeng2024ualign,wang2023retrosynthesis}. Specifically, models evaluated on mapped data benefit from explicit "shortcuts" provided by atom correspondence signals, whereas models using unmapped data must infer these relationships from scratch. This creates an uneven comparison. Our work addresses these discrepancies by developing a unified framework that strategically integrates chemical priors, such as reaction logic, while establishing a fair evaluation protocol.

\section{Methodology}
\label{sec:methods}

\subsection{Overview}
Our methodological framework advances chemical reaction prediction by systematically addressing key limitations in current approaches. We posit that true progress requires moving beyond simple parameter scaling to develop models with deep chemical understanding. Our architecture integrates three core innovations: (1) a \textbf{Hierarchical Cognitive Curriculum} that progressively builds chemical knowledge from syntax to semantics; (2) a \textbf{Latent Cycle-Consistency} principle that enforces reversibility in chemical transformations; and (3) an \textbf{Atom-Map Permutation Invariance} mechanism that forces learning of topological correspondences rather than superficial patterns. Together, these components enable a compact model to outperform significantly larger models while maintaining computational efficiency for downstream applications.

\subsection{Reaction Representation}

In our framework, a chemical reaction $r$ is formalized as a transformation between molecular sets, consistent with our problem definition in Section~\ref{subsec:problem_formalization}. For a reaction with $m$ reactants and $n$ products, we represent the reactants $\mathbf{R}$ and products $\mathbf{P}$ as concatenated SMILES strings:

\begin{equation}
	\mathbf{R} = r_1.r_2.\ldots.r_m \quad >> \quad \mathbf{P} = p_1.p_2.\ldots.p_n
\end{equation}

where $r_i$ and $p_j$ are SMILES strings of individual molecules, and "." denotes molecular separation.

\textbf{Atom-Atom Mapping (AAM)} provides crucial information by establishing explicit correspondences between reactant and product atoms. In AAM-annotated SMILES, each atom receives a unique identifier $\text{id} \in \{1, \dots, K\}$ where $K$ is the total number of mapped atoms in the reaction. The annotated representations are:

\begin{align*}
	\mathbf{R}_{\text{map}} &= [\text{Atom}_a:\text{id}_a][\text{Atom}_b:\text{id}_b]\ldots \\
	&>> \mathbf{P}_{\text{map}} = [\text{Atom}_x:\text{id}_x][\text{Atom}_y:\text{id}_y]\ldots
\end{align*}
Atoms sharing the same $\text{id}$ in $\mathbf{R}_{\text{map}}$ and $\mathbf{P}_{\text{map}}$ are mapped to each other. This establishes a mapping function $m: \mathcal{V}_R \rightarrow \mathcal{V}_P$, where $\mathcal{V}_R$ and $\mathcal{V}_P$ are the atom sets of reactants and products respectively.

The mapping identifiers create a ground-truth correspondence that makes the reaction transformation \textit{traceable}. For instance, in the following example:
\begin{verbatim}
	...[C:6]([O:5][C:2] ... [CH3:4])=[O:7].
	[CH3:8][C:9]=[O:10] ... [nH:15]...
	>> [CH3:1][C:2] ... [O:7][n:15]1 ... [cH:19] ... 
\end{verbatim}
The mapping $\{1, 2, \ldots, 19\}$ shows how atoms move through the reaction: atom \texttt{1} (methyl carbon) maintains its identity, while atom \texttt{15} transforms from \texttt{[nH:15]} to \texttt{[n:15]}, indicating bond formation.
 Models can exploit the specific numerical assignments ($\{1, 2, \ldots, K\}$) as shortcuts rather than learning the underlying chemical relationships. To address this, we introduce \textbf{Atom-Map Permutation Invariance (AMPI)} in Stage 3 of our curriculum (Section~\ref{subsec:curriculum}). 

Formally, let $\mathcal{I} = \{1, \dots, K\}$ be the set of mapping identifiers in a reaction. Given any permutation $\pi: \mathcal{I} \rightarrow \mathcal{I}$, the permuted mapping $\pi(\text{id})$ should yield chemically equivalent information. That is, the model should learn the \textit{relational topology}—which atoms correspond—rather than the specific numerical assignments.

This insight connects directly to our learning objectives: while AAM provides valuable training signal, AMPI ensures the model develops genuine chemical understanding rather than memorizing mapping patterns.

\subsection{Problem Formalization}
\label{subsec:problem_formalization}

Let $\mathcal{M}$ be the space of all valid molecules represented as  SMILES strings. A chemical reaction $r$ is a transformation from reactants $\mathbf{R} \in \mathcal{M}^*$ to products $\mathbf{P} \in \mathcal{M}^*$. We formulate three complementary prediction tasks that constitute a complete reaction reasoning cycle:

\begin{definition}[Reaction Prediction Tasks]
	Given a Transformer-based language model $f_\theta$ with parameters $\theta$, we define:
	\begin{align}
		\mathcal{T}_f &: \mathbf{R} \rightarrow \mathbf{P} &\text{(Forward Prediction)} \\
		\mathcal{T}_r &: \mathbf{P} \rightarrow \mathbf{R} &\text{(Retrosynthesis)} \\
		\mathcal{T}_{r|c} &: (\mathbf{P}, c) \rightarrow \mathbf{R} &\text{(Class-conditioned Retrosynthesis)}
	\end{align}
	where $c \in \mathcal{C}$ represents a reaction class label from a finite set $\mathcal{C}$.
\end{definition}

These tasks are inherently related through chemical reversibility principles. The \textbf{Cycle-Consistency Constraint} formalizes this relationship:

\begin{definition}[Chemical Cycle-Consistency]
	For an ideal model $f_\theta^*$, the composition of forward and backward predictions should approximate identity:
	\begin{equation}
		f_\theta^{(r)} \circ f_\theta^{(f)}(\mathbf{R}) \approx \mathbf{R} \quad \text{and} \quad f_\theta^{(f)} \circ f_\theta^{(r)}(\mathbf{P}) \approx \mathbf{P}
	\end{equation}
	where $f_\theta^{(f)}$ and $f_\theta^{(r)}$ denote the forward and retrosynthesis functions respectively.
\end{definition}

This constraint imposes strong regularization by ensuring the model learns transformations that respect chemical feasibility rather than arbitrary string mappings.

\subsection{Hierarchical Cognitive Curriculum}
\label{subsec:curriculum}

Our training paradigm follows a three-stage curriculum that mimics the cognitive progression from surface patterns to deep understanding.

\subsubsection{Stage 1: Syntactic Phase}
The initial stage establishes mastery over chemical syntax. Let $\mathcal{D}_{\text{syn}} = \{(\mathbf{R}_i, \mathbf{P}_i)\}_{i=1}^N$ be a dataset of canonical SMILES pairs. We optimize the standard language modeling objective:
\begin{equation}
	\mathcal{L}_{\text{syn}}(\theta) = -\frac{1}{N}\sum_{i=1}^N \sum_{t=1}^{T_i} \log P_\theta(y_t^{(i)} | y_{<t}^{(i)}, \mathbf{x}^{(i)})
\end{equation}
where $\mathbf{x}^{(i)}$ is the input sequence (e.g., reactants for $\mathcal{T}_f$), $y_t^{(i)}$ is the $t$-th token of the target sequence, and $T_i$ is the sequence length.

This stage builds fundamental competencies in: (1) SMILES grammar and syntax; (2) basic molecular graph-to-sequence translation; and (3) statistical patterns of common functional group transformations.

\subsubsection{Stage 2: Denoising Phase}
To develop robust chemical reasoning that transcends surface patterns, we introduce structured noise. For each input sequence $\mathbf{x}$, we define a corruption function $\mathcal{N}(\cdot)$ that applies one of:
\begin{itemize}[leftmargin=*]
	\item \textbf{Token Masking}: Randomly mask $k\%$ of tokens with a special [MASK] token.
	\item \textbf{Token Deletion}: Randomly delete $k\%$ of tokens.
\end{itemize}

The denoising objective becomes:
\begin{equation}
	\mathcal{L}_{\text{denoise}}(\theta) = -\mathbb{E}_{\mathbf{x} \sim \mathcal{D}}\left[\log P_\theta(\mathbf{x} | \mathcal{N}(\mathbf{x}))\right]
\end{equation}

This stage teaches the model to: (1) recover molecular identity from partial information; (2) recognize chemically implausible sequences; and (3) develop invariant representations across different SMILES linearizations.

\subsubsection{Stage 3: Semantic Phase with AMPI}
The final stage introduces explicit reaction mechanism information through Atom-Atom Mapping (AAM). Let $\mathbf{x}_{\text{AAM}}$ be an atom-mapped SMILES where atoms involved in bond changes are tagged with indices. Crucially, we implement \textbf{Atom-Map Permutation Invariance (AMPI)} to prevent shortcut learning.

\begin{definition}[AMPI Transformation]
	Given an atom-mapped SMILES $\mathbf{x}_{\text{AAM}}$ with $n$ mapped atoms having indices $\{a_1, \dots, a_n\}$, let $\pi: \{1,\dots,n\} \rightarrow \{1,\dots,n\}$ be a random permutation. The AMPI-transformed input $\mathbf{x}_{\text{AMPI}}$ is obtained by replacing each index $a_i$ with $\pi(a_i)$.
\end{definition}

The semantic learning objective with AMPI is:
\begin{equation}
	\mathcal{L}_{\text{sem}}(\theta) = -\mathbb{E}_{(\mathbf{x},\mathbf{y}) \sim \mathcal{D}_{\text{AAM}}, \pi \sim \Pi}\left[\log P_\theta(\pi^{-1}(\mathbf{y}) | \pi(\mathbf{x}))\right]
\end{equation}
where $\pi(\mathbf{x})$ applies the permutation to indices in the input, and $\pi^{-1}(\mathbf{y})$ reverses it in the output. This ensures the model learns \textit{which atoms correspond} rather than \textit{what numbers they have}.

\subsubsection{Parameter-Efficient Fine-Tuning} 
We employ Low-Rank Adaptation (LoRA) for training the LLMs:
\begin{equation}
	\Delta W = BA^T \quad \text{where} \quad B \in \mathbb{R}^{d \times r}, A \in \mathbb{R}^{r \times k}, r \ll \min(d,k)
\end{equation}
This approach ensures that the knowledge acquired during different phase is preserved while allowing for efficient adaptation to different stages.

\subsection{Plan-Based Reasoning as Latent Variable Model}
\label{subsec:plan_reasoning}

To enable explicit, step-by-step chemical reasoning, we formulate prediction as a latent variable model. Let $z$ be a latent plan variable representing the reasoning steps. The generation process is:
\begin{align}
	z &\sim P_\theta(z | \mathbf{x}) \\
	\mathbf{y} &\sim P_\theta(\mathbf{y} | \mathbf{x}, z)
\end{align}

We employ a deterministic posterior approximation $Q(z|\mathbf{x},\mathbf{y}) = \delta(z - z^*)$ where $z^*$ is the ground-truth reasoning plan (automatically extracted from reaction mechanisms or generated by rule-based systems). The evidence lower bound (ELBO) becomes:
\begin{equation}
	\log P_\theta(\mathbf{y}|\mathbf{x}) \geq \underbrace{\log P_\theta(\mathbf{y}|\mathbf{x}, z^*)}_{\text{reconstruction}} + \underbrace{\log P_\theta(z^*|\mathbf{x})}_{\text{plan prediction}} - \underbrace{D_{\text{KL}}(\delta(z-z^*) \| P_\theta(z|\mathbf{x}))}_{\text{regularization}}
\end{equation}

In practice, we implement this by training the model to generate sequences formatted as:
\[
\langle\text{input}\rangle \mathbf{x} \langle\text{plan}\rangle z^* \langle/\text{plan}\rangle \langle\text{answer}\rangle \mathbf{y} \langle/\text{answer}\rangle
\]

The plan $z^*$ includes key mechanistic steps: (1) identification of reaction centers; (2) electron movement patterns; (3) bond formation/breakage events; and (4) stereochemical considerations where applicable.

\begin{proposition}[Plan-Guided Generalization]
	Let $\mathcal{R}$ be the space of chemical reactions and $\mathcal{Z}$ the space of valid mechanistic plans. If there exists a surjective mapping $\phi: \mathcal{Z} \rightarrow \mathcal{R}$ such that reactions with similar mechanisms have nearby representations in $\mathcal{Z}$, then the plan-based model $f_\theta(\mathbf{x}) = \arg\max_\mathbf{y} P_\theta(\mathbf{y}|\mathbf{x}, z)$ with $z \sim P_\theta(z|\mathbf{x})$ has lower sample complexity than direct prediction models.
\end{proposition}

The detail plan-based tokens in the prompt are shown in Appendix~\ref{appendix:prompt}.
\subsection{Integrated Training Algorithm}
\label{subsec:training_algorithm}

The complete training procedure integrates all components:

\begin{algorithm}[H]
	\caption{Hierarchical Curriculum Training}
	\begin{algorithmic}[1]
		\STATE Initialize model parameters $\theta$
		\STATE \textbf{Stage 1: Syntactic Pre-training}
		\FOR{epoch = 1 to $E_1$}
		\STATE Sample batch from $\mathcal{D}_{\text{syn}}$ with equal task probability
		\STATE Update $\theta$ using $\nabla_\theta \mathcal{L}_{\text{syn}}$
		\ENDFOR
		\STATE \textbf{Stage 2: Denoising Pre-training}
		\FOR{epoch = 1 to $E_2$}
		\STATE For each example, apply corruption $\mathcal{N}(\cdot)$ with probability $p_{\text{noise}}$
		\STATE Update $\theta$ using $\nabla_\theta \mathcal{L}_{\text{denoise}}$
		\ENDFOR
		\STATE \textbf{Stage 3: Semantic Fine-tuning}
		\FOR{epoch = 1 to $E_3$}
		\STATE For each AAM example, apply random permutation $\pi \sim \Pi$
		\STATE Update $\theta$ using combined gradient:
		\[
		\nabla_\theta \left( \mathcal{L}_{\text{sem}} + \lambda_{\text{plan}} \mathcal{L}_{\text{plan}} \right)
		\]
		\ENDFOR
	\end{algorithmic}
\end{algorithm}

The detail prompts are shown in Appendix~\ref{appendix:prompt}.

\section{Experiment}
\subsection{Experiment settings}
\paragraph{Baselines}
For the template-based methods: RetroSim~\citep{coley2017computer}, NeuralSym~\citep{segler2017neural}, GLN~\citep{dai2019retrosynthesis}, LocalRetro~\citep{chen2021deep}, and RetroKNN~\citep{xie2023retrosynthesis}. On the semitemplate-based methods: G2G~\citep{shi2020graph}, RetroXpert~\citep{yan2020retroxpert}, RetroPrime~\citep{wang2021retroprime}, GraphRetro~\citep{somnath2020learning}, and Graph2Edits~\citep{zhong2023retrosynthesis}. For the template-free methods: Seq2Seq~\citep{schwaller2018found}, SCROP~\citep{zheng2019predicting}, AutoSynRoute~\citep{lin2020automatic}, GET~\citep{mao2021molecular}, DMP fusion~\citep{zhu2023dual}, Tied Transformer~\citep{kim2021valid}, MEGAN~\citep{sacha2021molecule}, Augmented Transformer~\citep{tetko2020state}, GTA~\citep{seo2021gta}, Graph2SMILES (D-GCN)~\citep{tu2022permutation}, Retroformer~\citep{wan2022retroformer}, Chemformer~\citep{irwin2022chemformer}, Ualign~\citep{zeng2024ualign}, R-SMILES~\citep{zhong2022root}, RetroExplainer~\citep{wang2023retrosynthesis}, EditRetro~\citep{han2024retrosynthesis}, RXNGraphformer~\citep{xu2025unified}.  Also the LLM-based models: RetroDFM~\citep{zhang2025reasoning}, ChemDual~\citep{lin2025enhancing}.

\paragraph{Datasets}
We evaluate our model on the widely used USPTO (United States Patent and Trademark Office) benchmark datasets, which are standard in retrosynthesis research ~\citep{lowe2012extraction,dai2019retrosynthesis}.. The USPTO dataset contains chemical reactions extracted from US patents, covering a broad spectrum of organic synthesis. To assess the model's performance across different tasks and scale, we utilize three key variants:USPTO-50K~\citep{schneider2016s}: A curated subset of 50,000 high-quality reactions, commonly used for benchmarking due to its manageable size and high data quality.USPTO-480K (MIT)~\citep{jin2017predicting}: A larger subset containing approximately 480,000 reactions, providing a more extensive training set for models requiring greater data diversity. USPTO-FULL~\citep{dai2019retrosynthesis}: The complete dataset of approximately one million reactions, representing the most comprehensive benchmark for large-scale retrosynthesis prediction.The detailed dataset statistics are shown in Table~\ref{tab:datasets}.

\paragraph{Implementation Details}
As shown in Table~\ref{tab:implementation_details},we implement our framework using the Qwen2.5-0.5B architecture as our base model.  We use the AdamW optimizer with a learning rate of $1\times10^{-4}$, $\beta_1 = 0.9$, $\beta_2 = 0.999$, and weight decay of 0.01. Training employs a linear learning rate scheduler with warmup over 10\% of the total training steps.
For the LoRA components, we use a rank $r = 64$ and $\alpha = 128$, applying adapters to the query, key, value, and output projections in attention layers, as well as the gate and up/down projections in feed-forward layers. 
All training can be down within 24GB memory GPU.

\paragraph{Evaluation}
To ensure fair and comprehensive evaluation, we rigorously separate performance on atom-mapped and unmapped test sets. For each task $\mathcal{T}_k$, we evaluate:
$$
\text{Acc}@k = \frac{1}{|\mathcal{D}_{\text{test}}|} \sum_{(x,y) \in \mathcal{D}_{\text{test}}} \mathbb{I} \left[ \hat{y} \in \text{top-}k(y) \right]$$
where $\hat{y}$ is the model's prediction and $\mathbb{I}$ is the indicator function. We report separate results for mapped ($\mathcal{D}_{\text{test}}^{\text{mapped}}$) and unmapped ($\mathcal{D}_{\text{test}}^{\text{unmapped}}$) evaluation sets to thoroughly assess generalization capability across different data formats.

\begin{table}[!h]
	\centering
	\vspace{-5pt}
	\caption{Dataset Information}
	\label{tab:datasets}
	\begin{tabular}{l r r r}
		\toprule
		Dataset & Train Size & Validation Size & Test Size \\
		\midrule
		USPTO\_480k & 409,035 & 30,000 & 40,000 \\
		USPTO\_50k & 40,008 & 5,001 & 5,007 \\
		USPTO\_full & 810,496 & 101,311 & 101,311 \\
		\bottomrule
	\end{tabular}
\end{table}

\subsection{Implementation Details}

Our framework is implemented using the Qwen2.5-0.5B architecture as the foundation model. The Qwen2.5-0.5B model provides an optimal balance between computational efficiency and performance, featuring 0.5 billion parameters with a hidden dimension of 2,048, 16 attention heads, and 24 transformer layers. This architecture supports the extended sequence lengths required for chemical reaction prediction while maintaining manageable memory requirements.

We employ the AdamW optimizer with a learning rate of $1\times10^{-4}$, $\beta_1 = 0.9$, $\beta_2 = 0.999$, and weight decay of 0.01. Training utilizes a linear learning rate scheduler with warmup over 10\% of the total training steps. The model processes sequences with a maximum length of 4,096 tokens to accommodate extended SMILES representations and plan-based reasoning steps.

\begin{table*}[!t]
	\centering
	\caption{Retrosynthesis Results on USPTO-50k with and without Reaction Type. The blue boxes are the best in each section.}
	\label{tab:uspto50k_results}
	\vspace{-9pt}
	\small
	\begin{tabular}{l l cccc cccc ccc}
		\toprule
		\multirow{2}{*}{\bf Category} & 
		\multirow{2}{*}{\bf Methods} & 
		\multicolumn{4}{c}{\bf Without Reaction Type} & 
		\multicolumn{4}{c}{\bf With Reaction Type} & 
		\multicolumn{3}{c}{\bf Features} \\
		\cmidrule(lr){3-6} \cmidrule(lr){7-10} \cmidrule(lr){11-13}
		& & Top-1 & Top-3 & Top-5 & Top-10 & Top-1 & Top-3 & Top-5 & Top-10 & \makecell{\scriptsize Templ.} & \makecell{\scriptsize Map} & \makecell{\scriptsize Test\\Aug.} \\
		\midrule
		
		\multirow{5}{*}{
			\rotatebox[origin=c]{90}{Template-based}
		}& 
		RetroSim& 37.3 & 54.7 & 63.3 & 74.1 & 52.9 & 73.8 & 81.2 & 88.1 & $\checkmark$ & - & - \\
		& NeuralSym & 44.4 & 65.3 & 72.4 & 78.9 & 55.3 & 76.0 & 81.4 & 85.1 & $\checkmark$ & - & - \\
		& GLN & 52.5 & 69.0 & 75.6 & 83.7 & 64.2 & 79.1 & 85.2 & 90.0 & $\checkmark$ & $\checkmark$ & $\times$ \\
		& LocalRetro & 53.4 & 77.5 & 85.9 & 92.4 & 63.9 & 86.8 & 92.4 & 96.3 & $\checkmark$ & $\checkmark$ & $\times$ \\
		& \cellcolor{blue!10}RetroKNN & \cellcolor{blue!10}57.2 & \cellcolor{blue!10}78.9 & \cellcolor{blue!10}86.4 & \cellcolor{blue!10}92.7 & \cellcolor{blue!10}66.7 & \cellcolor{blue!10}88.2 & \cellcolor{blue!10}93.6 & \cellcolor{blue!10}96.6 & $\checkmark$ & - & - \\
		\midrule
		\multirow{5}{*}{
			\rotatebox[origin=c]{90}{Semitemplate}
		}
		& 
		G2G & 48.9 & 67.6 & 72.5 & 75.5 & 61.0 & 81.3 & 86.0 & 88.7 & $\times$ & $\checkmark$ & $\times$ \\
		& RetroXpert & 50.4 & 61.1 & 62.3 & 63.4 & 62.1 & 75.8 & 78.5 & 80.9 & $\times$ & $\checkmark$ & $\checkmark$ \\
		& RetroPrime & 51.4 & 70.8 & 74.0 & 76.1 & 64.8 & 81.6 & 85.0 & 86.9 & $\times$ & $\checkmark$ & $\checkmark$ \\
		& GraphRetro & 53.7 & 68.3 & 72.2 & 75.5 & 63.9 & 81.5 & 85.2 & 88.1 & $\times$ & $\checkmark$ & $\times$ \\
		& \cellcolor{blue!10}Graph2Edits & \cellcolor{blue!10}55.1 & \cellcolor{blue!10}77.3 & \cellcolor{blue!10}83.4 & \cellcolor{blue!10}89.4 & \cellcolor{blue!10}67.1 & \cellcolor{blue!10}87.5 & \cellcolor{blue!10}91.5 & \cellcolor{blue!10}93.8 & $\checkmark$ & $\times$ & - \\
		\midrule
		\multirow{16}{*}{
			\rotatebox[origin=c]{90}{Template-free}
		}
		& 
		Seq2Seq & 37.4 & 52.4 & 57.0 & 61.7 & 37.4 & 52.4 & 57.0 & 61.7 & $\times$ & - & - \\
		& SCROP & 43.7 & 60.0 & 65.2 & 68.7 & 59.0 & 74.8 & 78.1 & 81.1 & $\times$ & $\times$ & $\times$ \\
		& AutoSynRoute & 43.1 & 64.6 & 71.8 & 78.7 & - & - & - & - & $\times$ & $\times$ & $\times$ \\
		& GET & 44.9 & 58.8 & 62.4 & 65.9 & - & - & - & - & $\times$ & $\times$ & $\times$ \\
		& DMP fusion & 46.1 & 65.2 & 70.4 & 74.3 & - & - & - & - & $\times$ & $\times$ & $\times$ \\
		& Tied Transformer & 47.1 & 67.2 & 73.5 & 78.5 & - & - & - & - & $\times$ & $\times$ & $\times$ \\
		& MEGAN & 48.1 & 70.7 & 78.4 & 86.1 & 60.7 & 82.0 & 87.5 & 91.6 & $\times$ & $\checkmark$ & $\times$ \\
		& Augmented Transformer & 48.3 & - & 73.4 & 77.4 & - & - & - & - & $\times$ & $\times$ & $\checkmark$ \\
		& GTA & 51.1 & 67.6 & 74.8 & 81.6 & - & - & - & - & $\times$ & $\checkmark$ & $\checkmark$ \\
		& Graph2SMILES (D-GCN) & 52.9 & 66.5 & 70.0 & 72.9 & - & - & - & - & $\times$ & $\times$ & $\times$ \\
		& Retroformer & 53.2 & 71.1 & 76.6 & 82.1 & 64.0 & 82.5 & 86.7 & 90.2 & $\times$ & - & - \\
		& Chemformer & 54.3 & - & 62.3 & 63.0 & - & - & - & - & $\times$ & $\times$ & $\checkmark$ \\
		& Ualign & 53.5 & 77.3 & 84.6 & 90.5 & 66.4 & 86.7 & 91.5 & 95.0 & $\times$ & $\times$ & $\checkmark$ \\
		& R-SMILES & 56.3 & 79.2 & 86.2 & 91.0 & - & - & - & - & $\times$ & $\times$ & $\checkmark$ \\
		& \cellcolor{blue!10}RetroExplainer & 57.7 & 79.2 & 84.8 & 91.4 & \cellcolor{blue!10}66.8 & \cellcolor{blue!10}88.0 & \cellcolor{blue!10}92.5 & \cellcolor{blue!10}95.8 & $\times$ & - & $\checkmark$ \\
		& \cellcolor{blue!10}EditRetro & \cellcolor{blue!10}60.8 & \cellcolor{blue!10}80.6 & \cellcolor{blue!10}86.0 & \cellcolor{blue!10}90.3 & - & - & - & - & $\times$ & $\checkmark$ & $\checkmark$ \\
		\midrule
		
		\multirow{4}{*}{
			\rotatebox[origin=c]{90}{LLM-based}
		} & ChemDual-8B & 50.0 &67.7 &70.5 &78.3 & - & - & - & - & $\times$ & $\times$ & $\times$\\
	& RetroDFM-R-7B & 59.0 & - & - & - & - & - & - & - & $\times$ & $\checkmark$ & $\times$ \\
		&  {RxnNano-0.5B(w/oAAM)} & \underline{69.8} & \underline{89.0} & \underline{92.8} & \underline{95.8} & \underline{72.0} & \underline{89.7} & \underline{93.5} & \underline{96.0} & $\times$ & $\times$ & $\times$ \\
		&\textbf{RxnNano-0.5B(AAM)} & \textbf{75.1} & \textbf{90.8} & \textbf{93.4} & \textbf{96.6} & \textbf{75.7} & \textbf{91.2} & \textbf{94.9} & \textbf{96.8} & $\times$ & $\checkmark$ & $\times$ \\
		\bottomrule
	\end{tabular}
\end{table*}

For parameter-efficient fine-tuning, we implement Low-Rank Adaptation (LoRA) with rank $r = 64$ and scaling parameter $\alpha = 128$. LoRA adapters are applied to the query, key, value, and output projections in attention layers, as well as the gate, up, and down projections in feed-forward layers. This configuration enables efficient adaptation while adding only 0.2\% additional parameters to the base model.

The complete training process requires approximately 72 hours on a single A30 GPU with 24GB memory. The combination of Qwen2.5-0.5B's compact architecture with LoRA fine-tuning ensures that our approach remains accessible while achieving state-of-the-art performance on chemical reaction prediction tasks.

\begin{table*}[htbp]
	\centering
	\setlength{\tabcolsep}{9pt} 
	\caption{Retrosynthesis and Forward Prediction Results on USPTO-full and USPTO-480k Datasets}
	\vspace{-5pt}
	\label{tab:uspto_full_480k}
	\begin{tabular}{lccccc @{\hspace{2em}} ccccc}
		\toprule
		& \multicolumn{5}{c}{\textbf{USPTO-full (retrosynthesis)}} & \multicolumn{5}{c}{\textbf{USPTO-480k (forward prediction)}} \\
		\cmidrule(lr){2-6} \cmidrule(lr){7-11}
		Methods & Top-1 & Top-3 & Top-5 & Top-10 & & Top-1 & Top-3 & Top-5 & Top-10 \\
		\midrule
		Transformer baseline & 42.9 & -- & -- & 66.8 & & -- & -- & -- & -- \\
		Molecular Transformer & -- & -- & -- & -- & & 88.6 & 93.5 & 94.2 & 94.9 \\
		MEGAN & -- & -- & -- & -- & & 86.3 & 92.4 & 94.0 & 95.4 \\
		Chemformer & -- & -- & -- & -- & & 91.3 & -- & 93.7 & 94.0 \\
		Graph2SMILES & 45.7 & -- & -- & 62.9 & & 90.3 & 94.0 & 94.6 & 95.3 \\
		RXNGraphformer & 47.4 & 63.0 & 67.4 & 71.6 & & 90.6 & 94.3 & 94.9 & 95.5 \\
		RetroDFM-R-7B & 50.5 & 67.6 & 72.7 & 77.5 & & - & - & - & - \\
		\textbf{RxnNano-0.5B(Ours)} & \textbf{62.1} & \textbf{66.9} & \textbf{71.9} & \textbf{76.8} & & \textbf{94.1} & \textbf{95.2} & \textbf{96.3} & \textbf{97.6} \\
		\bottomrule
	\end{tabular}
\end{table*}

\subsection{Main results}
\textbf{Performance on USPTO-50k}
The results in Table~\ref{tab:uspto50k_results} demonstrate that our method achieves state-of-the-art performance on the USPTO-50k benchmark. Our model obtains \textbf{75.1\%} top-1 accuracy in type-unknown and \textbf{75.7\%} in type-known scenarios, representing improvements of 23.5\% and 13.32\% over the best existing methods, respectively.
The minimal performance difference between type-unknown and type-known conditions (0.6 percentage points) indicates effective internalization of reaction type information during training. This contrasts with traditional methods that show significant degradation without type guidance.

\textbf{TTA and AAM}. Moreover, methods with {test time augmentation} often conduct 20 times augmentation in both training and testing phase to improve the performance. However, the model is evaluated on more synthetic dataset in this case and lead to unfair comparison. 
The introducing of AAM contain more reaction information but models with AAM does not consistently outperform others. This suggest training strategy is important even with high quality data.
Ours without AAM and test time augmentation still achieves sota performance compared with all the baselines.

\textbf{Scalability to Large-Scale Datasets and forward prediction}
Table~\ref{tab:uspto_full_480k} demonstrates our method's scalability. On USPTO-full (810k examples), we achieve \textbf{62.1\%} top-1 accuracy, outperforming RetroDFM-R-7B by +22.9\%. On USPTO-480k forward prediction, we obtain \textbf{94.2\%} top-1 accuracy and outperform all the baselines.
The consistent performance gains across dataset sizes indicate that our advantages scale with data volume. Ours enables effective knowledge transfer between retrosynthesis and forward prediction tasks, with each task informing and regularizing the other.

\begin{figure}
	\centering
	\includegraphics[width=0.95\linewidth]{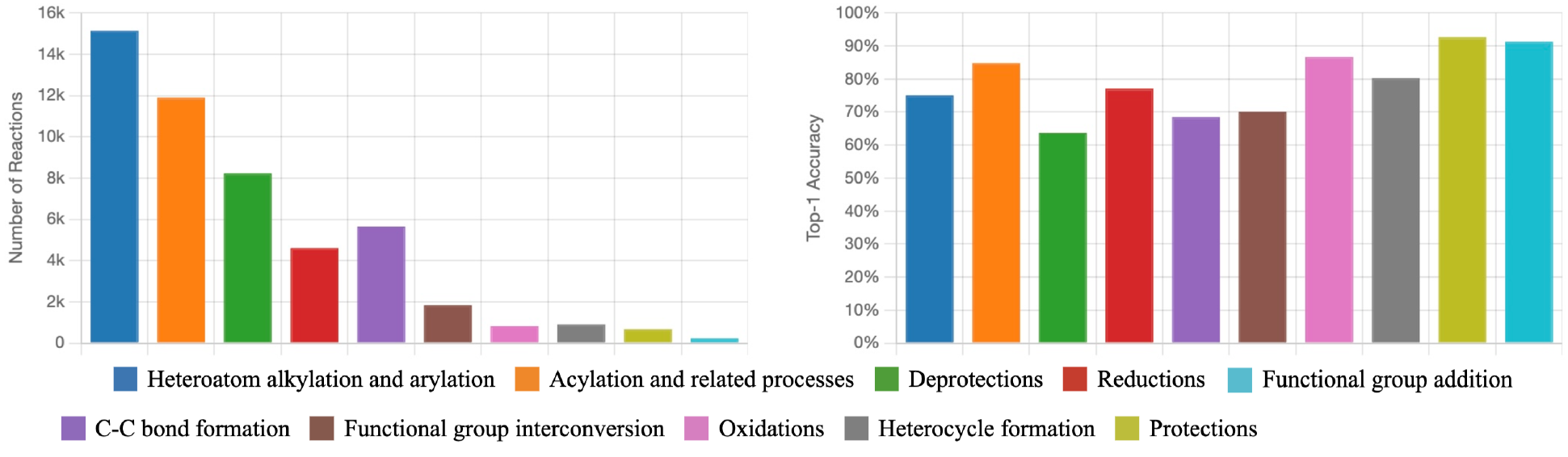}
	\vspace{-5pt}
	\caption{
		Reaction types in USPTO-50K (left) and corresponding model performance metrics (right)}
	\label{fig:performance_class}
	\vspace{-5pt}
\end{figure}

\textbf{Reaction Type Generalization Analysis}.
Figure~\ref{fig:performance_class} reveals consistent performance across reaction types, with particular strength in complex transformations involving multiple bond changes. Although the datasets is unbalanced across different class. Traditional methods struggle with these due to template combinatorial explosion or invalid intermediate generation.

\begin{table}[htbp]
	\centering
	\small
		\setlength{\tabcolsep}{1pt}
	\caption{Performance comparison of large language models on chemical tasks}
	\vspace{-5pt}
	\label{tab:llm-comparison}
	\begin{tabular}{lccccc}
		\toprule
		\textbf{Model} & \textbf{Size} & \textbf{Acc (\%)} & \textbf{ExtraData} & \textbf{Tokenizer} \\
		\midrule
		\multicolumn{5}{c}{\textbf{Base LLM}} \\
		\midrule
		GPT-4o & -- & 0.7 & No & No \\
		GPT-4.1-mini & -- & 11.3 & No & No \\
		DeepSeek-V3 & 671B & 8.6 & No & No \\
		o1-mini & -- & 12.2 & No & No \\
		DeepSeek-R1 & 671B & 11.2 & No & No \\
		Qwen2.5-0.5B & 0.5B & 0.0 & No & No \\
		\midrule
		\multicolumn{5}{c}{\textbf{Finetuned LLM}} \\
		\midrule
		ChemDFM-v1.5-13B & 13B & 17.9 & Chemical & No \\
		RetroDFM-R-7B & 7B & 59.0 & Chemical & No \\
		ChemDual (Llama-3.1-8B) & 8B & 49.95 & No & Modified \\
		Ours-0.5B (without AAM) & 0.5B & 69.8 & No & No \\
		Ours-0.5B (with AAM) & 0.5B & \textbf{75.1} & No & No \\
		\bottomrule
	\end{tabular}
\end{table}

\subsection{Foundation Models vs. Specialized Approaches}
\label{subsubsec:llm_comparison}

Table~\ref{tab:llm-comparison} compares our approach with foundation and fine-tuned LLMs.Part of the LLMs performance are adopted from previous papers~\cite{zhang2025reasoning}.  Larger models with more data might not inherently perform better. This suggest that current LLMs can't perform well on these tasks. Direct fine-tuning with larger amount of data and modification of the tokenizer with chemical knowledge can not provide superior performance. Even massive models (e.g., DeepSeek-V3, 671B) achieve only 8.6–12.2\% accuracy, demonstrating that general capabilities do not transfer directly to chemical reasoning. Fine-tuned LLMs (e.g., RetroDFM-R, 7B) improve to 59.0\% but remain far below our 0.5B model (75.1\% with AAM, 69.8\% without). Crucially, our model achieves this superior performance without extra data or specialized tokenizers, indicating that strategic design focused on chemical understanding, rather than scale alone, yields more efficient and effective solutions for specialized domains like reaction prediction.

\subsection{Ablation Study}
\label{subsec:ablation}
\begin{table}[t]
	\centering
	\caption{Ablation study results on USPTO-50k. Results are reported in Top-1 accuracy (\%).}
	\label{tab:ablation_study}
	\small
	\begin{tabular}{lcc}
		\toprule
		\textbf{Method Variant} & \textbf{Unknown Type} & \textbf{Known Type} \\
		\midrule
		\textbf{Ours (AAM)} & \textbf{75.1} & \textbf{75.7} \\
		\quad - cycle & 71.8 & 73.3 \\
		\quad - plan & 74.1 & 74.6 \\
		\quad - syntax & 72.7 & 73.5 \\
		\quad - denoising & 71.4 & 72.8 \\
		\quad - semantic & 53.8 & 55.3 \\
		\midrule
		\textbf{Ours (w/o AAM)} & 69.8 & 72.0 \\
		\quad - syntax & 68.4 & 70.5 \\
		\quad - denoising & 67.3 & 69.6 \\
		\quad - AMPI & 34.5 & 36.8 \\
		\bottomrule
	\end{tabular}
\end{table}
 We conduct ablation studies on the USPTO-50k dataset as shown in Table~\ref{tab:ablation_study}.
\textbf{Efficacy of Curriculum Components.} The three-stage curriculum proves essential for robust chemical reasoning. Removing the three stage causes performance drops.  SMILES grammar mastery facilitates subsequent learning. The \textit{denoising pre-training} stage contributes to developing robust representations. Most critically, the \textit{semantic fine-tuning} stage (which includes AMPI for the w/o AAM variant) shows the largest impact. This confirms that explicit reaction mechanism learning, enforced through structured objectives, is indispensable for genuine chemical understanding rather than surface pattern matching.

\textbf{Chemical Cycle-Consistency.} The cycle-consistency constraint provides consistent benefits across both configurations. By enforcing that forward and backward predictions approximate identity compositions, this constraint acts as a powerful regularizer that encourages chemically feasible transformations.

\textbf{Critical Role of AMPI for Generalization.} In the w/o AAM setting, removing AMPI causes the  performance drop. This validates our hypothesis that without explicit atom-mapping supervision, models tend to memorize specific numerical patterns rather than learning relational correspondences. AMPI forces the model to develop permutation-invariant representations of atomic relationships, enabling generalization beyond the specific indexing schemes seen during training.

\begin{figure}[htbp]
	\centering
	\begin{minipage}{0.35\textwidth}
		\centering
		\includegraphics[width=\linewidth]{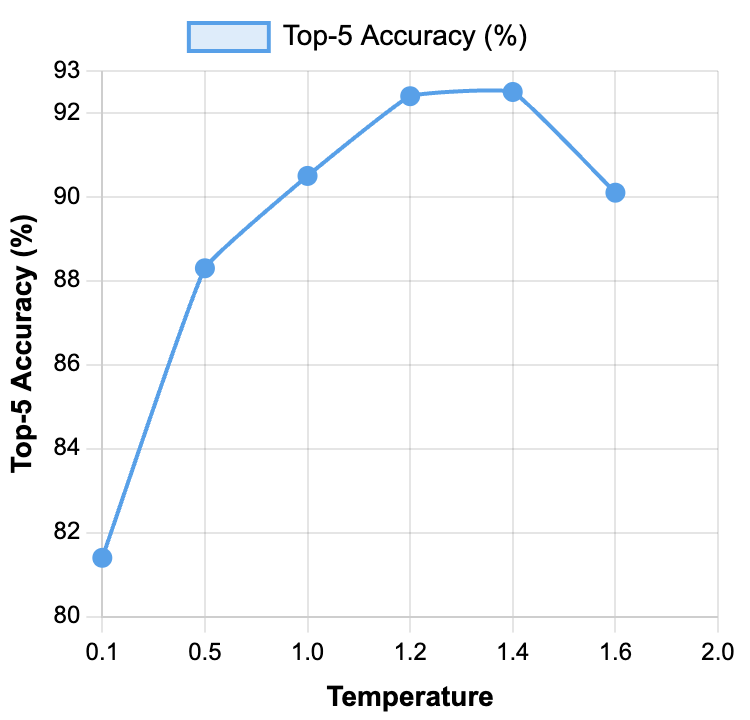}
		\caption{Top-5 accuracy v.s temperature.}
		\label{fig:topk_temperature}
	\end{minipage}
\end{figure}

\textbf{Plan-based reasoning}
improve the performance. The planning mechanism reduce the generation uncertainty with fixed domain tokens.

\textbf{Temperature sensitivity}
Figure~\ref{fig:topk_temperature} presents a temperature sensitivity analysis. For top-1 accuracy, we employ greedy decoding (temperature = 0), which yields optimal performance for single-prediction scenarios. However, top-k accuracy improves with increasing temperature, reaching a maximum at T=1.6 before declining. Excessive randomness (T $>$ 1.6) degrades performance.

\section{Limitations and Future Directions}

While demonstrating strong performance and efficiency, we trained the base model over single step chemical reaction prediction. For complex multi-step reactions need further study. 
Though we show how good a compact a model can achieve, it is still valuable to train a larger model with more diverse data.
Future work will focus on enhanced planning mechanisms for complex pathways and use  the base model as agent for downstream tasks. Incorporation of practical constraints (feasibility, cost, safety) represents another important direction.

\clearpage
\newpage


\bibliographystyle{ACM-Reference-Format}
\bibliography{sample-base}

\appendix
\label{appendix}

\section{Prompt formulation} 
\label{appendix:prompt}
Here is the simplified prompt template for the tasks. The full version is in~\ref{prompt:task_full}.
\begin{tcolorbox}[
	colback=gray!10, 
	colframe=gray!80, 
	title=System Prompt,
	fonttitle=\bfseries,
	boxrule=0.3mm,
	sharp corners,
	width=\columnwidth
	]
	\noindent Respond in the following format:\\
	\texttt{<plan>}\\
	...\\
	\texttt{</plan>}\\
	\texttt{<answer>}\\
	Provide the final answer in JSON format as specified in the instruction.\\
	\texttt{</answer>}
	\vspace{-4pt}
	\label{prompt:system}
\end{tcolorbox}

Example of the LLM prompt formation:
\begin{tcolorbox}[
	colback=gray!10, 
	colframe=gray!80, 
	title=Retrosynthesis Prompt template,
	fonttitle=\bfseries,
	boxrule=0.3mm,
	sharp corners,
	width=\columnwidth
	]
	Task: Retrosynthesis
	
	Given the product SMILES: "{product}"
	
	Predict the reactants required to synthesize this product.
	
	\#\#\# Instruction:\\
	...\\
	- Return the predicted reactants in SMILES format as a JSON object:\\
	{{"reactants": "SMILES\_string"}}.
	\vspace{-4pt}
	\label{prompt:task}
\end{tcolorbox}

Example of plan tokens:
\begin{tcolorbox}[
	colback=gray!10, 
	colframe=gray!80, 
	title= \texttt{<plan>} for retrosynthesis,
	fonttitle=\bfseries,
	boxrule=0.3mm,
	sharp corners,
	width=\columnwidth
	]
	\noindent \texttt{</plan>}\\
	To predict the reactants for the product SMILES:\\
	1. Identify key functional groups and structural features in the product.\\
	2. Propose retrosynthetic disconnections based on common reaction types (e.g., esterification, amide formation, sulfonamide formation, heterocycle synthesis).\\
	3. Validate that the proposed reactants are chemically feasible and can form the product under standard conditions.\\
	\texttt{</plan>}\\
	\vspace{-4pt}
	\label{prompt:plan}
\end{tcolorbox}

\subsection{Detail prompt formulation}

\begin{tcolorbox}[
	colback=gray!10, 
	colframe=gray!80, 
	title=Retrosynthesis Prompt,
	fonttitle=\bfseries,
	boxrule=0.3mm,
	sharp corners,
	width=\columnwidth
	]
	Task: Retrosynthesis
	
	Given the product SMILES: "{product}"
	
	Predict the reactants required to synthesize this product.
	
	\#\#\# Instruction:
	- Think step-by-step to identify the reactants based on the product SMILES.\\
	- Consider common retrosynthetic disconnections and reaction types (e.g., amide formation, esterification, nucleophilic substitution).\\
	- Ensure the SMILES string is valid, includes atom mapping if present in the product, and uses '.' to separate multiple reactants.\\
	- Note: This is an unmapped SMILES representation. If no atom mapping is provided or if atom mapping numbers are all 0 or -1, treat it as unmapped. Mapped and unmapped representations are similar but differ in format, with mapped including explicit atom correspondences. Still includes the atom mapping in the predicted reactants.\\
	- Return the predicted reactants in SMILES format as a JSON object:\\
	{{"reactants": "SMILES\_string"}}.
	\vspace{-4pt}
	\label{prompt:task_full}
\end{tcolorbox}

\begin{tcolorbox}[
	colback=gray!10, 
	colframe=gray!80, 
	title=Retrosynthesis with reaction type known Prompt,
	fonttitle=\bfseries,
	boxrule=0.3mm,
	sharp corners,
	width=\columnwidth
	]
	Task: Retrosynthesis
	
	Given the product SMILES: "{product}"and reaction class: \"{rxn\_class}
	
	Predict the reactants required to synthesize this product.
	
	\#\#\# Instruction:
	- Think step-by-step to identify the reactants based on the product SMILES.\\
	- Consider common retrosynthetic disconnections and reaction types (e.g., amide formation, esterification, nucleophilic substitution).\\
	- Ensure the SMILES string is valid, includes atom mapping if present in the product, and uses '.' to separate multiple reactants.\\
	- Note: This is an unmapped SMILES representation. If no atom mapping is provided or if atom mapping numbers are all 0 or -1, treat it as unmapped. Mapped and unmapped representations are similar but differ in format, with mapped including explicit atom correspondences. Still includes the atom mapping in the predicted reactants.\\
	- Return the predicted reactants in SMILES format as a JSON object:\\
	{{"reactants": "SMILES\_string"}}.
	\vspace{-4pt}
	\label{prompt:task_class}
\end{tcolorbox}

\begin{tcolorbox}[
	colback=gray!10, 
	colframe=gray!80, 
	title=Forward prediction Prompt,
	fonttitle=\bfseries,
	boxrule=0.3mm,
	sharp corners,
	width=\columnwidth
	]
	Task: Forward prediction
	
	Given the reactants SMILES: "{reactants}""
	
	Predict the product of this reaction.
	
	\#\#\# Instruction:\\
	- Think step-by-step to identify the product based on the reactants SMILES.\\
	- Consider common reaction types (e.g., amide formation, esterification, nucleophilic substitution).\\
	- Ensure the SMILES string is valid, includes atom mapping if present.\\
	- Return the predicted product in SMILES format as a JSON object:\\
	{{"product": "SMILES\_string"}}.
	"""
	\vspace{-4pt}
	\label{prompt:task_forward}
\end{tcolorbox}

\section{Implementations}
It is shown in Table~\ref{tab:implementation_details}.
\begin{table}[htbp]
	\centering
	\caption{Model Architecture and Training Configuration}
	\label{tab:implementation_details}
	\begin{tabular}{lc}
		\toprule
		\textbf{Parameter} & \textbf{Value} \\
		\midrule
		\textbf{Base Model} & Qwen2.5-0.5B \\
		Hidden dimension & 2,048 \\
		Attention heads & 16 \\
		Transformer layers & 24 \\
		Maximum sequence length & 4,096 \\
		\midrule
		\textbf{Optimization} & \\
		Optimizer & AdamW \\
		Learning rate & $1\times10^{-4}$ \\
		$\beta_1$, $\beta_2$ & 0.9, 0.999 \\
		Weight decay & 0.01 \\
		Batch size & 32 \\
		Warmup steps & 5,000 \\
		Total training steps & 50,000 \\
		\midrule
		\textbf{LoRA Configuration} & \\
		Rank ($r$) & 64 \\
		Alpha ($\alpha$) & 128 \\
		Dropout & 0.1 \\
		\midrule
		\textbf{Hardware} & \\
		Device & A30 GPU \\
		GPU Memory & 24GB \\
		\bottomrule
	\end{tabular}
\end{table}


\end{document}